\documentclass[letterpaper, 10 pt, conference]{ieeeconf}  

\usepackage{array}
\usepackage{longtable}
\usepackage{caption}
\usepackage{graphicx}
\usepackage{amsmath}
\usepackage{subfig}
\newcolumntype{P}[1]{>{\centering\arraybackslash}p{#1}}
\newcolumntype{M}[1]{>{\centering\arraybackslash}m{#1}}

\newcounter{eqn}

\makeatletter
\newcommand{\putindeepbox}[2][0.7\baselineskip]{{%
    \setbox0=\hbox{#2}%
    \setbox0=\vbox{\noindent\hsize=\wd0\unhbox0}
    \@tempdima=\dp0
    \advance\@tempdima by \ht0
    \advance\@tempdima by -#1\relax
    \dp0=\@tempdima
    \ht0=#1\relax
    \box0
}}
\makeatother

\IEEEoverridecommandlockouts                              

\title{\LARGE \bf
Autodetection and Classification of Hidden Cultural City Districts from Yelp Reviews
}

\author{Harini Suresh$^{1}$ and Nick Locascio$^{2}$
\thanks{*This work was supported by Joshua Tenenbaum}
\thanks{$^{1}$H. Suresh is an Undergraduate in Computer Science,
        Massachussets Insititude of Technology, Cambridge, MA 02139, USA
        {\tt\small hsuresh at mit.edu}}%
\thanks{$^{2}$N. Locascio is an Undergraduate in Computer Science,
        Massachussets Insititude of Technology, Cambridge, MA 02139, USA
        {\tt\small njl at mit.edu}}%
}

\begin{document}

\maketitle
\thispagestyle{empty}
\pagestyle{empty}

\begin{abstract}

Topic models are a way to discover underlying themes in an otherwise unstructured collection of documents. In this study, we specifically used the Latent Dirichlet Allocation (LDA) topic model on a dataset of Yelp reviews to classify restaurants based off of their reviews.  Furthermore, we hypothesize that within a city, restaurants can be grouped into similar ``clusters” based on both location and similarity.  We used several different clustering methods, including K-means Clustering and a Probabilistic Mixture Model, in order to uncover and classify districts, both well-known and hidden (i.e. cultural areas like Chinatown or hearsay like ``the best street for Italian restaurants") within a city. We use these models to display and label different clusters on a map.  We also introduce a topic similarity heatmap that displays the similarity distribution in a city to a new restaurant.

\end{abstract}

\section{INTRODUCTION}
Cities are often defined by their domineering cultural characteristics. However, particular cities themselves harbor a large variety of different cultural districts. Streets separated by just a few blocks may give very different impressions. These implicit boundaries and classifications are not documented on official maps, and usually are only learned with much time and experience living in a particular city.\\

\subsection{Motivation}
We believe that having a sense of these districts is valuable to a much wider population.  Some examples are:\\ 
1. New Businesses: For business-owner or entrepreneur looking to open a new restaurant or expand to a different location, knowing which areas of a city harbor restaurants very similar to or different than that particular business is doubtless a valuable insight.  \\
2. Newcomers: For tourists, people moving in, or anyone else new to the city, it is often an arduous and daunting task to get a sense of things such as where they are most likely to find a good Thai restaurant, \emph{the} block to go for Dim Sum, or the best area for a dressy, upscale dinner with good wine.  \\
3. Anyone looking to explore: Even people who have already have a sense of the city can be surprised by a hole-in-the-wall cafe or undiscovered area.  The LDA model we describe can identify the most highly weighted classification for a particular area as well as secondary classifications.  This allows it to uncover more hidden characteristics of a particular area besides the most highly weighted.  This is interesting information in itself, but can also be used as a backbone of a recommendation system. If a  person really enjoys a particular area of town, this model could discover and rank other non-obvious areas that share similar traits.\\

From a cognitive science point of view, we think trying to model these questions is an interesting experiment to test the accuracy of methods like LDA and Probabilistic Mixture Models to model human cognition. Recent cognitive science research has had major successes in probabilistic generative models of human cognition [12, 13]. Specifically, research by Tenenbaum shows strong support for Bayesian concept learning [14] and Sanborn et. al. use Dirichlet Process Mixture Models for category learning that emulates human learning [15]. Using techniques like these in this paper, we try to recreate the kind of map a local might build up in their head over time of the different subsections of their city. \\ 

\subsection{The Yelp Dataset}
The Yelp Academic dataset was released in 2013 and has grown to include over 42,000 businesses with over 1 million reviews [9]. The dataset has been used in academic papers for sentiment analysis, word layout systems, and recommendation engines, among other research areas. The quality and sheer size of the dataset is of high value to our research and its natural language user reviews are pivotal to our cultural detection and classification system.

\section{METHODS}

\subsection{The LDA Model}
Latent Dirichlet Allocation (LDA), first introduced by Blei et. al. in 2003 [1], has been applied to numerous and diverse fields: from computer vision [2,3] to recommendation systems [4] to spam filtering [5]. LDA hypothesizes that a collection of documents $D$ can be treated as a "bag of words" where each document d is generated by the following process, given hyperparameters $\alpha$ and $\beta$:\\
\begin{enumerate}
\item[1.]Assume each topic $k$ has a fixed distribution over all words in D that is $\phi_{k} \sim Dirichlet(\beta)$
\item[2.]Choose the document's topic distribution $\theta_{d} \sim Dirichlet(\alpha)$
\item[3.]To generate each word w:\\
  a. Choose a topic $z_{i}$ from $Discrete(\theta_{d})$ \\
    b. Choose a word $w_{i}$ from $Discrete(\phi_{z_{i}})$ \\
\end{enumerate}

Using this model, LDA is able to learn the topic mixtures, $p(z | d)$, for the documents on which it is trained, in an unsupervised manner. \\

\subsection{Training LDA}
To implement LDA, we used tools from the Python Library Gensim, which provides functionality to analyze semantic structure in texts [6].  Based off of the results of the Expectation Maximization algorithm used by Huang et. al. [7] to determine the optimal number of topics for Yelp restaurant reviews in Phoenix, we chose $K$ = 50 as the number of topics to extract. We used hyperparameters $\alpha$ and $\beta$ with symmetric 1.0/K priors. \\

We cleaned the reviews to remove punctuation, numbers, and a list of stopwords made up of the ``English Stop Words" list in the Scikit-learn python library [11]. Additionally, we specified that after this initial cleaning, the model should only consider the 40,000 median frequency words. This eliminated words that only appeared a handful of times, as well as generic food-related words that appeared many times. These words provide little information gain and removing them dramatically increased the convergence time of our LDA training.\\

We trained the model on all restaurant reviews (around 1.1 million) from Las Vegas. Training uses the online inference algorithm described by Hoffman et. al. [8] and results in an LDA topic model object that can be queried with new, unseen documents to return an optimal topic distribution. We used our model to predict topic weights for each restaurant. In addition, the model contains the static word distributions $\phi_{k}$ for each topic.

\subsection{Training Examples}
Our LDA model produced 50 topics. Each topic is a collection of word-weight couples. Words with high corresponding weight values are most representative of the topic. The topic word weights are normalized such that $\sum\limits_{w_i}^W w_i = 1$. Table I is a small sampling of selected topics our model generated. Table I displays the topic \#, the label we chose for the topic based on its word distribution, and the word distribution. The full list of topics and their weights can be viewed in the Appendix.\\

\begin{table}[h]
\caption{Selected Topic Assignments}
\label{Topic Assignment Table}
\begin{center}
\begin{tabular}{| M{0.6cm} | M{2cm} | M{3.4cm} |}
\hline
Topic \# & Label & Words Distribution\\
\hline
2 & Mexican Food & 0.043*tacos + 0.037*taco + 0.026*asada + 0.024*carne + 0.024*mexican + 0.019*burrito + 0.010*salsa + 0.009*fries + 0.009*beans + 0.008*roberto's \\ \hline
7 & Night Club & 0.013*music + 0.012*fun + 0.009*club + 0.007*cool + 0.006*party + 0.006*lounge + 0.005*group + 0.005*floor + 0.004*dance + 0.004*girls \\ \hline 
45 & Casino & 0.027*hotel + 0.022*casino + 0.020*room + 0.010*stay + 0.009*downtown + 0.006*staying + 0.006*pool + 0.005*street + 0.005*stayed + 0.005*fremont \\ \hline
\end{tabular}
\end{center}
\end{table}

\begin{table}[h]
\caption{Selected Topic Prediction Results}
\label{Topic Table}
\begin{center}
\begin{tabular}{| M{1cm} | M{2cm} | M{3cm} |}
\hline
Restaurant & Top 2 Topic \#'s & Top Words from Top Topic\\
\hline
Pho Vietnam Restaurant & 4, 15 & pho, vietnamese, rolls, broth\\
\hline
Myxx Hookah Lounge & 7, 15 & music, fun, club, cool, party\\
\hline
Romano's Macaroni Grill & 24, 30 & pasta, italian, bread, server\\
\hline
\end{tabular}
\end{center}
\end{table}

\subsection{LDA Inference}

We used our trained LDA model to predict topic distributions for each of the 3855 restaurants. A restaurant's topic distribution is a collection of coupled topic numbers and corresponding weights. Topics with high corresponding weight values are most representative of the restaurant. These topic distribution weights are normalized such that $\sum\limits_{w_i}^W w_i=1$. A sampling of topic predictions is shown in Table II.

\subsection{Clustering Preparation}
We tested several clustering methods in order to group restaurants into appropriate clusters. We assume that culinary districts in a city are characterized by closeness and similarity of restaurants. In our model, therefore, we represent each restaurant as a combined vector of its coordinate position and its LDA assigned topic weight distribution. This vector has 52 dimensions, 2 of which represent the spatial location of the restaurant, and 50 of which represent the restaurant's LDA topic weights. \\

\noindent1. Scaling Procedure \\

Since the spatial coordinates and topic weights are measured in different spaces, their values are on different scales. To prevent our results being arbitrarily skewed by these different units of scale, we used a scaling procedure, multiplying the topic weight distributions by a constant $S$.\\

By varying $S$, we can give the topic weights more or less influence over the clustering. When $S=0$, the clustering is equivalent to clustering based only on location. As S increases, topic weights are given more control over the clustering. When $S>>coordinates$, the clustering is done purely by topic similarity.\\

Our goal was to find an $S$ such that close-together clusters of restaurants would be grouped into a single cluster, and points on the outer edges of these clusters would identify themselves with the cluster that best matched their topic distribution. In this way, we allow for a Chinese restaurant to 'escape' a nearby cluster of Italian restaurants. \\

Since we are not using pure spatial features, our clustering may result in some clusters overlapping and interweaving. This allows our model to be representative of the real world of cultural mixing and fuzzy cultural boundaries.\\

To determine a reasonable scaling factor $S$, we constructed a plausible scenario. A Chinese Restaurant ($CR$) lies betwen two clusters - a primarily Italian cluster ($I$) and a primarily Chinese cluster ($C$). $CR$ is 0.25 mi from $I$'s center and 0.75 mi from $C$'s center.  We want to choose an S such that Dist($CR$,$C$) $<$ Dist($CR$, $I$): that is, we want the Chinese restaurant to be classified into the ``Chinese Restaurant" cluster despite it being closer in spatial coordinates to the Italian cluster's center. The $Dist$ function is a Euclidean distance metric between the two vectors. Let's assume $C$ and $CR$ share the same topic distributions, and that $CR$ and $I$ share zero topics in their distributions. We define $Dist(CR,C)$ as

\begin{equation}
Dist(R,Cluster) = \sqrt{(\Delta x)^2 + S*\sum_{t_j}^{T} (\Delta t_j)^2}
\end{equation}

So our goal is to find an S such that:

\begin{equation}
Dist(CR,C) < Dist(CR, I)
\end{equation}

\begin{gather}
\sqrt{(0.75)^2 + (0.0)^2} < \sqrt{(0.25)^2 + (S*1.0)^2 + (S*1.0)^2}\nonumber\\
0.75 < \sqrt{(0.25)^2 + 2S^2}\nonumber\\
0.5625 < 0.0625 + 2S^2\nonumber\\
0.5 < 2S^2\nonumber\\
0.25 < S^2\nonumber\\
\end{gather}

\begin{align}
S > 0.5
\end{align}

This calculation of S, however, assumes topic distributions are all-or-none, when in fact most restaurants are a mixture of a few topics. In fact, we determined the mean \# of topics assignments a restaurant received to be 5. We found a typical restaurant to have 1 dominating topic comprising of at least 0.5 of the weight and 4 subtopics comprising of the rest of the weight. We performed  a more advanced analysis of the same scenario and found $S > 0.913$. The analysis can be found in Equation 10 of the appendix. \\

\noindent2. Normalization \\

Since some topics are inherently more common than other topics due to the high prevalence of some restaurant types, we wanted to avoid our model becoming unfairly skewed by very common topics such as a ``pizza" topic. There are 355 pizza restaurants in Las Vegas, comprising of 9.2\% of all Las Vegas restaurants. To avoid the scenario where all clusters are labeled as ``pizza" simply because of the uniformly large number of these restaurants across all clusters, we vertically normalize the topic weights for each restaurant. We define $w$ to be a 50 dimensional topic weight vector of a restaurant, and $N$ to be the number of restaurants. We define 

\begin{align}
V\_Norm(w) = \frac{w^{i}}{W^{i}}
\end{align}

where 
\begin{align}
W^{i} = \sum\limits_{j=1}^{N} w_{j}^{i}
\end{align}

This normalization can be thought of as dividing out the 'background' of a city's restaurant distribution, ensuring clusters will be dominated by notable exceptions to the average: we don't want to point out that pizza restaurants are pretty much evenly distributed in high quantities all around Vegas, but rather discover when they, or another type of restaurant, are appear in \emph{notably} high quantities. We then horizontally re-normalize each topic vector so that the values remain at the same scale.\\

\noindent3. Determining the Number of Clusters \\

To determine the optimal number of clusters, we first used the 'Elbow Method', which looks at the percentage of variance explained as a function of the number of clusters. The idea is that we should choose a number of clusters such that adding more clusters doesn't significantly improve the modeling. We performed clustering with $C$ = 5 to $C$ = 35 clusters and plotted the variance quantity $W_{k}$ vs. $C$, where $W_{k}$ is the sum of the normalized intra-cluster sums of sqaures [16]. Figure 1 shows a plot of log($W_{k}$) vs. $C$.\\

\begin{figure}[h!]
  \caption{Elbow Graph}
  \centering
    \includegraphics[width=0.45\textwidth]{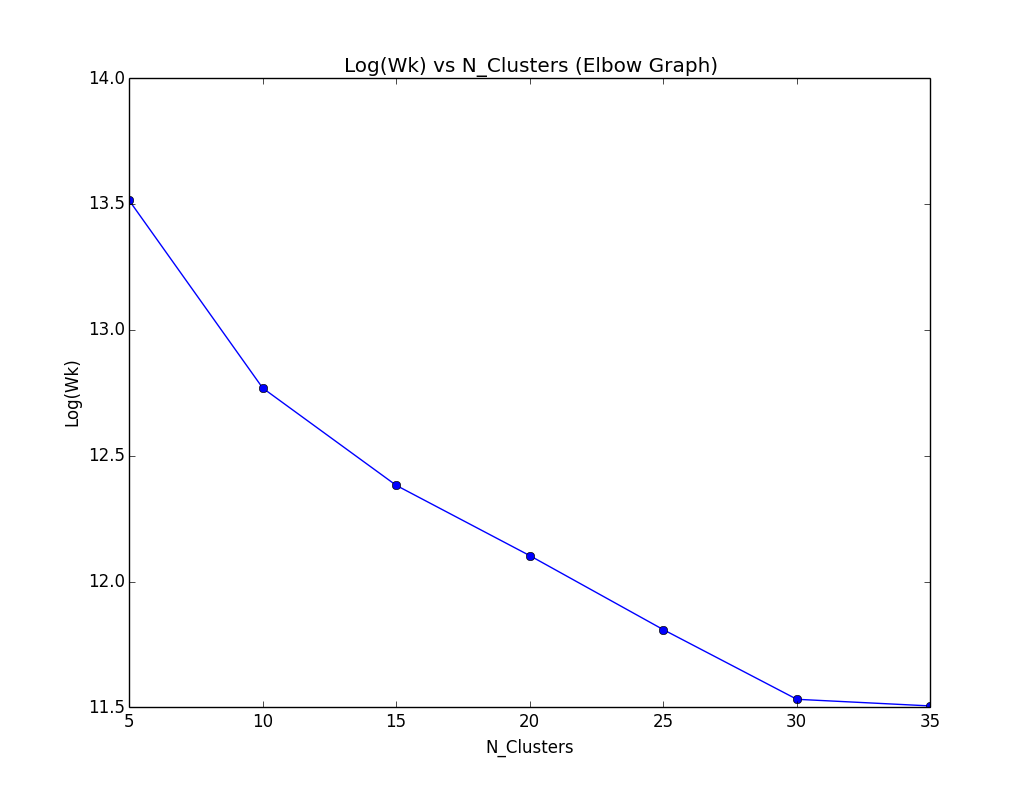}
\end{figure}

\begin{figure}[h!]
  \caption{Gap Statistic Graph}
  \centering
    \includegraphics[width=0.45\textwidth]{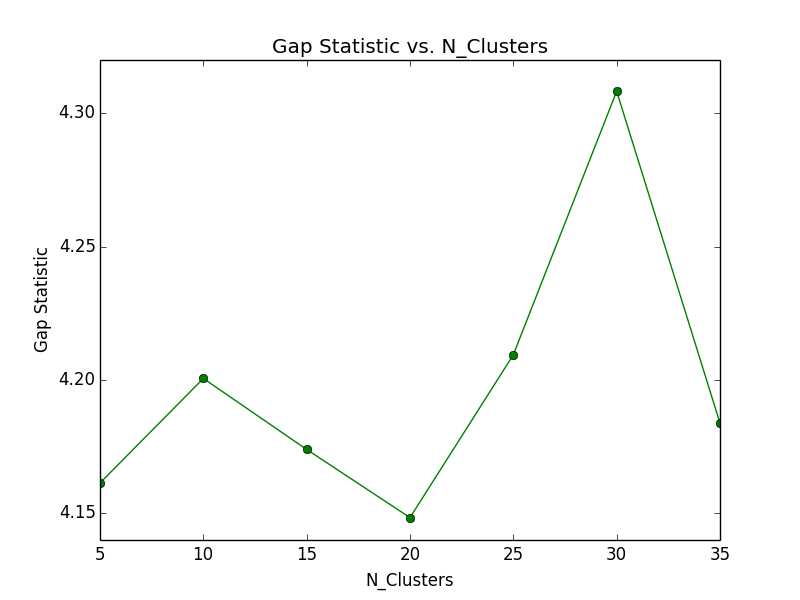}
\end{figure}

The Elbow Method involves visually choosing the 'elbow' of the graph where the slope changes most drastically. We determined our elbow happens at $C$ = 30. However, determining the elbow of a graph is not a well-defined process, and in fact this is one of the known weaknesses of the Elbow Method.\\

Because of the shortcomings of the elbow method, we also used the Gap Statistic [10, 16] to determine the optimal $C$ with which to cluster. The Gap Statistic is a way to to standardize the comparison of the ``variance explained" metric used in the Elbow Method. The Gap Statistic takes the approach of standardizing the variance explained against a null reference distribution of the data (distribution with no apparent underlying clusters). The Gap Statistic method involves calculating the difference between the variance explained for the dataset and the variance explained for the null reference distribution. This difference is known as the Gap Statistic. The $C$ value that yields the greatest Gap Statistic (greatest difference in variance) is the optimal $C$ value for clustering the data. Figure 2 shows the results of the Gap Statistic.\\

The Gap Statistic predicts that $C$=30 is the optimal number of clusters for our data. This confirms our identification of the 'elbow' was indeed correct.\\

\subsection{Clustering}
\noindent1. K-means Clustering \\

K-means clustering is a clustering algorithm that will find $C$ centroids to cluster a data set. The K-Means algorithm converges on a centroid distribution that minimizes the sum of squares of
distances between cluster centroids and the corresponding data points that are classified by them.\\

Using tools from the Python library Scikit-learn [11], we performed K-means clustering on all 3855 Vegas restaurants with random K++ means initialization and 300 iterations, specifying $C=30$ and $S=0.913$. We used a Euclidean distance metric for our clustering and classification of the 52-dimensional restaurant vectors. The result of this clustering can be seen in Figure 3.\\
\begin{figure}[h!]
  \caption{Las Vegas K-means Clusters}
  \centering
    \includegraphics[width=0.45\textwidth]{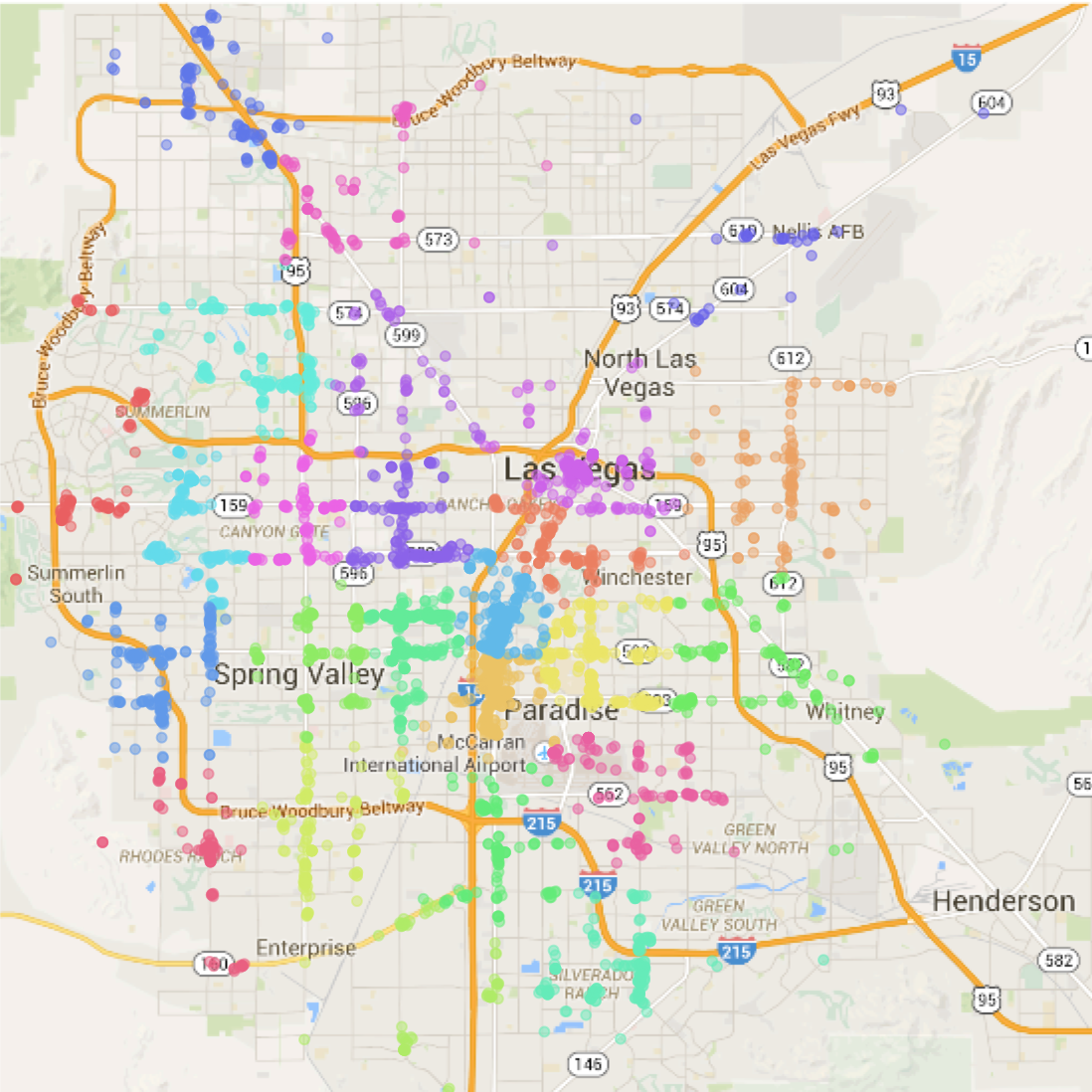}
\end{figure}

The clusters have a median of 148 members in each with a standard deviation of 80.\\

\noindent2. Gaussian Mixture Model \\

A Gaussian mixture model (GMM) is a probabilistic generative model that assumes all the data points are generated from a mixture of a finite number of Gaussian distributions. The GMM, in principle, is a weighted sum of $C$ component Gaussian densities. Each Gaussian distribution can be thought of as a cluster that can classify data points.\\

Also using tools from the Python library Scikit-learn, we trained our GMM with the Expectation Maximization Algorithm on 3855 restaurants specifying $C=30$ and $S=0.913$. The result of this clustering can be seen in Figure 3. The GMM clusters have a median of 200 members each, with mean 158 members and standard deviation of 40 members.\\

Most notably, GMM clusters varied from the K-means clusters in shape: the K-means clusters were nearly always spherical in shape, due to K-means minimizing distance. The Gaussian Mixture Model, however, is not limited to spherical clusters, as the Gaussian distributions that define its clusters are shaped by variances in each dimension. This results in some elongated elliptical clusters. \\

Some of the clusters consists of only restaurants that lie on a particular street. It may be that this behavior is actually beneficial. Oftentimes cultural districts within a city are highly street based, and the GMM model is flexible enough to detect clusters like this. The result of the GMM clustering can be seen in Figure 4. \\
\begin{figure}[h!]
  \caption{Las Vegas Gaussian Clusters}
  \centering
    \includegraphics[width=0.45\textwidth]{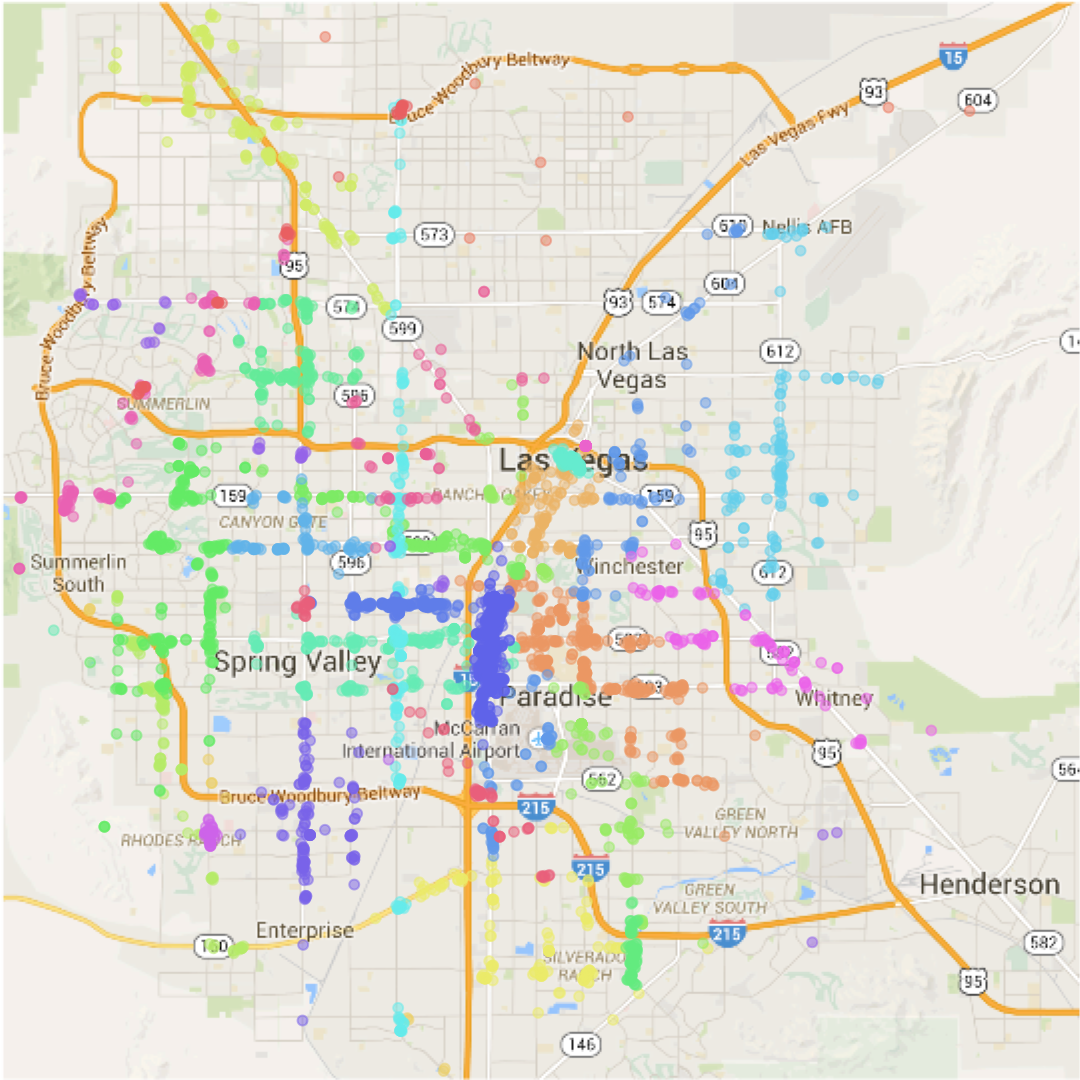}
\end{figure}
 
\subsection{Determining a Cluster's Label}
Once we determine relevant spatial and topical clusters, we are tasked with labeling the clusters. To determine the labeling of a cluster, we take the average topic vector for all restaurants in the cluster. We then chose the top \emph{two} topics that describe a cluster and use their human-attributed labels. These labels overlayed atop their cluster distributions are shown in Figure 5.\\

We chose to display the top two labels to uncover not only the most frequent topic within a cluster but also underlying categories which might be less obvious. \\

\begin{figure}[h!]
  \caption{Las Vegas K-means Clusters with Labelings}
  \centering
    \includegraphics[width=0.45\textwidth]{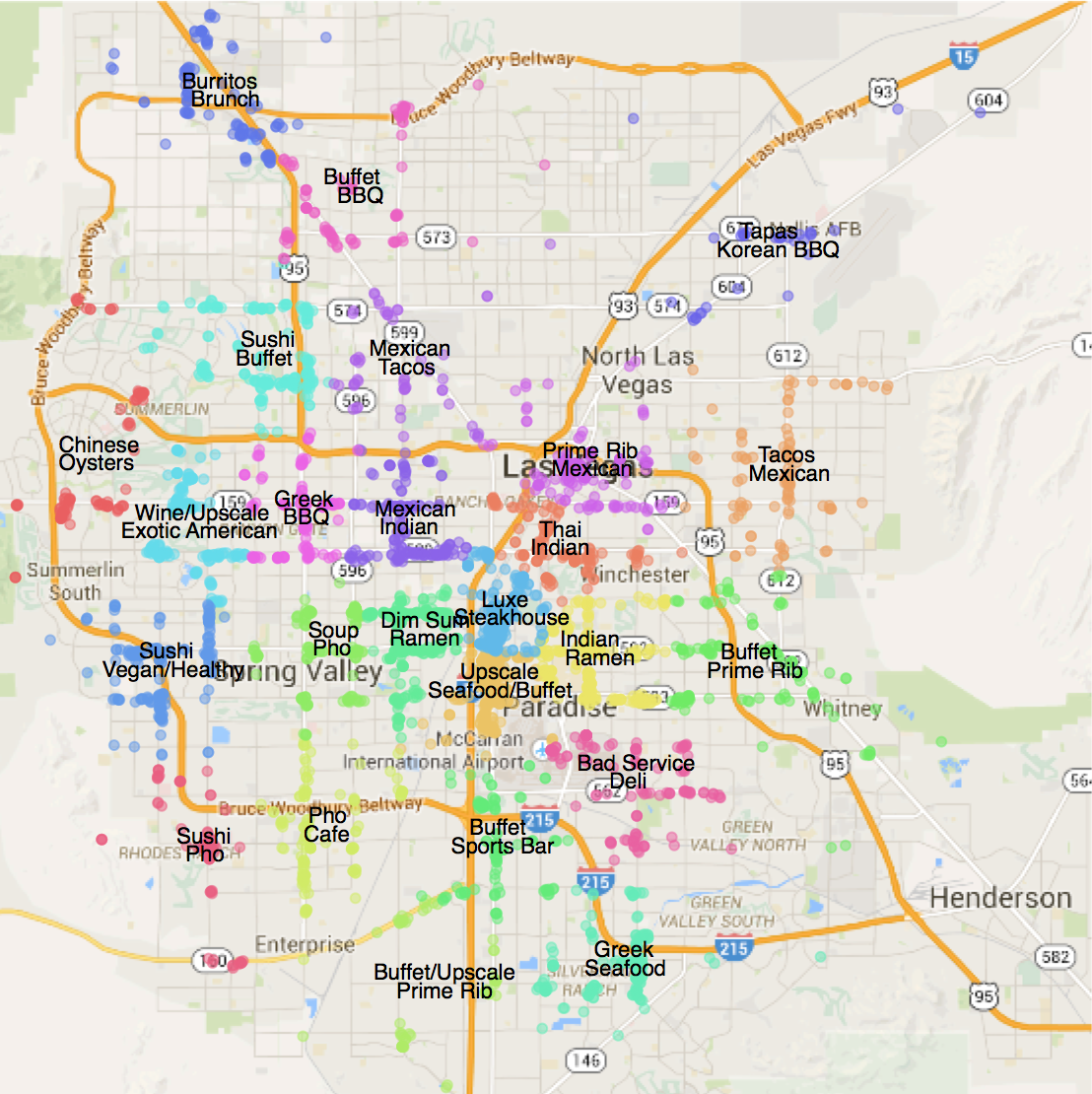}
\end{figure}

\begin{figure}[h!]
  \caption{Las Vegas Gaussian Clusters with Oriented Labelings}
  \centering
    \includegraphics[width=0.45\textwidth]{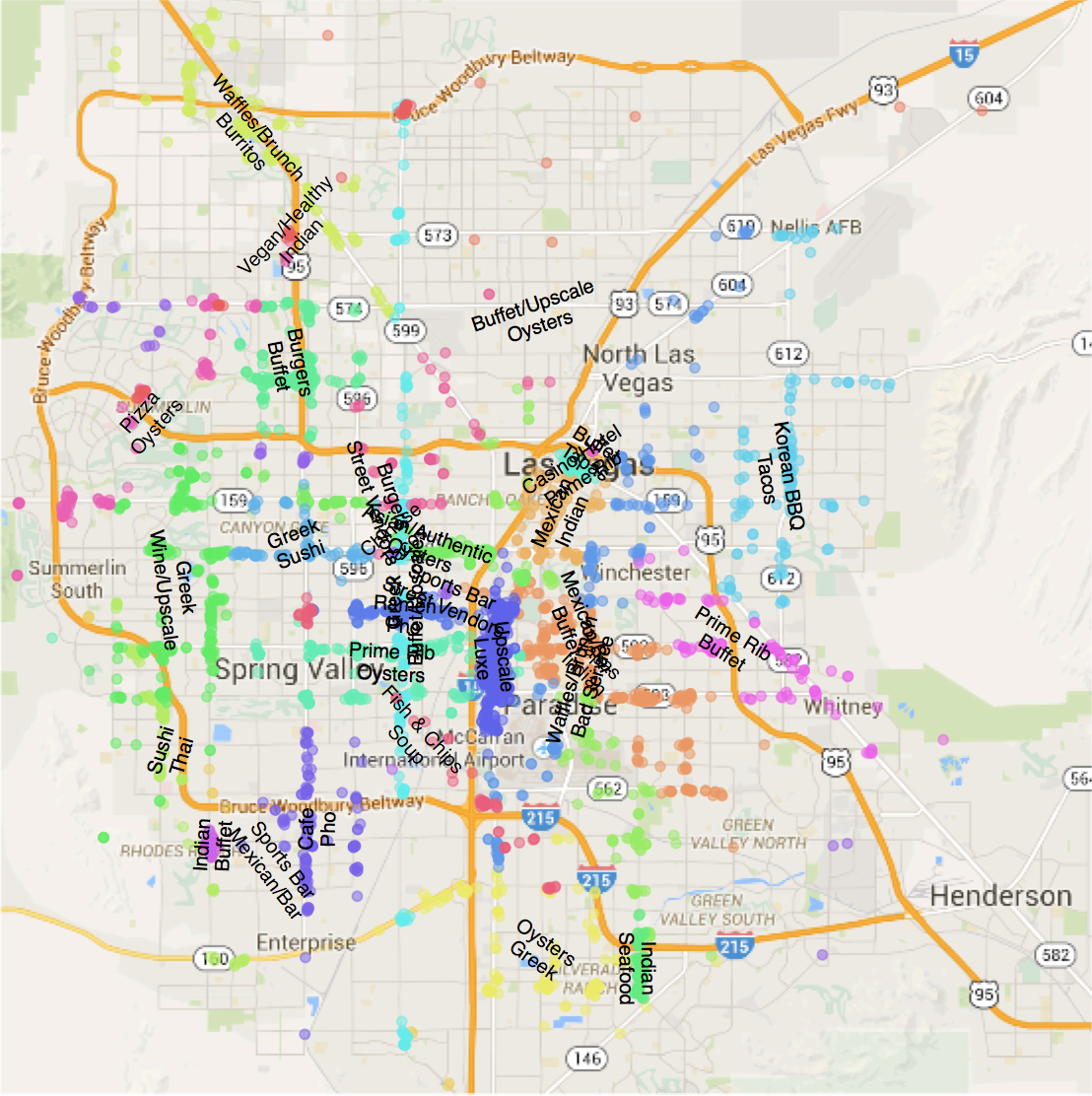}
\end{figure}

Using our Gaussian Mixture Clustering, we were able to enhance these labeling with appropriate orientations\footnote[3]{Unfortunately, the label rotations often result in the collision of labels. If this system were to be implemented as an analytics interface, this issue could be mitigated by a zoomable display, hiding of colliding labels, or other methods. Our paper, however, is concerned more with the exploration of these methods than the user interfacing and experience.}. Since each cluster is represented by a Gaussian with two dimensional variances, we are able to rotate the labelings to align with the direction of maximum Gaussian variance. These rotated labels have a tendency to orient with streets. The Waffles/Brunch label in the top left displays this rather useful property. These oriented labels overlayed atop their Gaussian cluster distributions are shown in Figure 6.\\

\subsection{Cultural HeatMap and Optimal Placement for New Restaurants}
While clustering restaurants on space and topics illuminates a city's many cultural centers, it does not show how a specific topic is distributed throughout the city. To show this distribution for a given topic we plotted topic similarity in a heatmap. We ran our LDA inference on a novel restaurant's reviews. From this we got a topic distribution of that novel restaurant. We divided the city into a a 20X20 grid of squares. For each square we calculated the average topic similarity from the center of the square to all restaurants in the city. We used a Gaussian Weight metric to scale topic similarity by proximity. For each square we calculated a similarity metric $Sim(center, novel)$ where:

\begin{multline}
Sim(center, novel) = \\ \frac{1}{\|\mathbf{R}\|} \sum_{r_i}^{R}2\sqrt{2} - dist(novel, r_i)*Gw(center, r_i)
\end{multline}
\\

where \\

\begin{description}
\item[$R$] = all restaurants in city
\item[$novel$] = the topic distribution of the novel restaurant
\item[$dist$] = euclidian distance metric for topic distributions
\end{description}

and

\begin{align}
Gw(square, restaurant) = G(x, \sigma, \mu)
\end{align}

where \\

\begin{description}
\item[$\mu$] $=$ the center of the square
\item[$\sigma$] $=\sqrt{2}*square\_width$
\item[$x$] $=$ the position of the restaurant
\end{description}

\begin{align}
G(x, \sigma, \mu) = \frac{1}{{\sigma \sqrt {2\pi } }}e^{{{ - \left( {x - \mu } \right)^2 } \mathord{\left/ {\vphantom {{ - \left( {x - \mu } \right)^2 } {2\sigma ^2 }}} \right. \kern-\nulldelimiterspace} {2\sigma ^2 }}}
\end{align}

Our similarity metric takes in to account topic similarity of the novel restaurant to each other restaurant in the city. We use a Gaussian weight to scale these topic similarities by distance. This allows restaurants near the square's center location to have most of the influence over the square's color. We calculated our similarity metric for every square in our grid and colored our Heat Map red for high values and blue for low values. The results of our heatmap generated by comparing the topics of the Restaurant "Pho Vietnamese Restaurant" to the restaurants of Vegas are shown in Figure 7.\\

\begin{figure}[h!]
  \caption{Pho Vietnamese Topic Similarity Heatmap}
  \centering
    \includegraphics[width=0.45\textwidth]{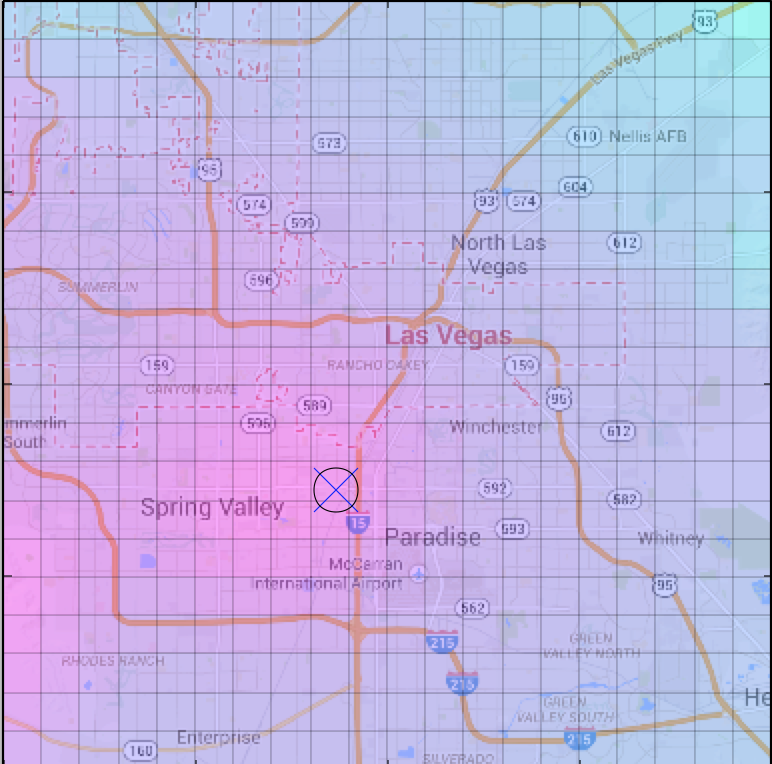}
\end{figure}

In Figure 8, the X indicates the actual location of the the restaurant (the similarity calculations were conducted without including this restaurant).  Our Heat Map shows that the restaurant is in an area of high topic similarity, which is accurate (Pho Vietnamese Restaurant is located in Las Vegas's Chinatown District).

\section{Discussion \& Applications}

\subsection{Evaluation of Results}
We found that the resultant LDA topics (Appendix Table III) were well-defined and descriptive. We observed that the words within a given topic fit well into a particular category of food type or culture, and we had very little trouble labeling them based off of the given words and weights. Additionally, the topics themselves seem reasonably distinct from each other with only few overlapping topics. The general area in which we saw the most overlap was the buffet restaurants topic. Topics \#0, \#5, \#48 each concerned buffet restaurants. However, looking at the words in each, we were able to distinguish ``Seafood/Buffet" (\#5) and ``Upscale/Buffet"(\#0) from a more general ``Buffet" topic (\#48). \\

Looking at the K-means clustering of Las Vegas restaurants, we observed that our clustering classifies areas defined beforehand: it put labels of ``Pho'' and ``Dim Sum'' on Chinatown, and ``Luxe'', ``Steakhouse'', ``Upscale'', and ``Seafood/Buffet'' over the Strip. Interestingly, it also split these clusters into smaller subclusters, for example separating a ``Dim Sum" and ``Ramen" cluster from a ``Pho" and ``Soup" cluster. This behavior may or may not be ideal: it may be identifying actual sub-districts, or future work may involve a final cluster merge step in which two clusters close in distance and topic similarity can be merged in to a single cluster.\\

The GMM also distinguished these already-known cultural areas. The shape and sizes of the clusters themselves were slightly more varied and often alligned with a particular street. This ability is interesting because oftentimes districts may be very street-based.\\

As all this learning was unsupervised, we are very interested in finding a metric to quantitatively determine accuracy across these different models.  One potential way to do this would be to conduct cognitive studies with people who live in or are familiar with particular cities. For example, we could compare our map with maps described by Las Vegas residents, or get a measure of how accurate they believe our map is. \\

\subsection{Applications of Automatic Cluster Labeling}
The automatic spatial and topical clustering and labeling approach outlined above is a general method that can be applied to any city. Figure 8 in the Appendix shows the results of labeling 2 other cities (Phoenix and Endinberg) with this method. These maps can be analytical tools with various applications including but not limited to determining new restaurant placement, understanding cultural regions of a city, discovering unexplored areas of one's city, choosing where to live, or what route to take on a stroll to the park.

\subsection{Applications of Topic Heat Map}
Like the automatic cultural labeling method, the Topic Heat Map can be used as a useful analytics tool. This map can be used to determine certain cultural hotbeds, both known and hidden. A hidden cultural hotbed may present a market opportunity for continued growth. The Topical Heat Map of a city may be an especially valuable asset to a new restaurant or chain looking to strategize where exactly to place a new store location. The Heat Map could actually be used to perform a detailed analysis on what kind of location, for different types of restaurants, is optimal (see IV part 3 for more detail).

\section{Further Research}
1) Using the timestamps on reviews, it is possible to filter reviews based on when they were written.  This would allow for creating dynamic maps using reviews within a moving a time window to see how culture changes: how new clusters emerge, split and merge.\\

2) In our study we use the Elbow Method and Gap Statistic to predetermine an appropriate number of clusters to use. Instead, it may prove valuable to use a Nested Chinese Restaurant Process to learn a hierarchy of clusters and subclusters. For example, this could split Chinatown into various subclusters under the general Chinese cluster. This could be used to label the graphs at various scales and zoom levels. Additionally, using a Chinese Restaurant Process as part of a Nonparametric Mixture Model would allow the model to flexibly add more clusters as needed, and may be more likely to find the optimal number of clusters.  \\

3) The similarity heatmap we developed along with Yelp star ratings could be used to analyze what kind of placement makes a restaurant successful.  For example, it may be that placing a restaurant right in the center of an area of very high similarity creates direct competition and comparison that is actually detrimental. At the same time, it may be that placing a restaurant in an area where it is completely out of place is also a bad idea.  A detailed analysis of where restaurants with varying star ratings fall on a similarity heatmap could provide valuable insight to businesses about what kind of placement is optimal. \\

4) Using Yelp user data and the classifications from this model, it is possible to create a recommendation system. Recommendations could be general areas or specific restaurant suggestions: for example, if a user likes several restaurants in a specific area/cluster, we imagine recommending to them another restaurant or area that shares similar topics.





\section*{ACKNOWLEDGMENT}

We would like to thank Joshua Tenenbaum for supporting this work. 


\onecolumn

\section*{APPENDIX}

\begin{center}
    \begin{longtable}{ | M{0.8cm} | M{2cm} | M{14cm} |}
    \caption{Topics Discovered By LDAModel}\\
\hline
\textbf{Topic \#} & \textbf{Human Label} & \textbf{Word Distribution} \\
\hline
\endfirsthead
\multicolumn{3}{c}%
{\tablename\ \thetable\ -- \textit{Continued from previous page}} \\
\hline
\textbf{Topic \#} & \textbf{Human Label} & \textbf{Word Distribution} \\
\hline
\endhead
\multicolumn{3}{r}{\textit{Continued on next page}} \\
\endfoot
\hline
\endlastfoot
    \hline
0 & Buffet/Upscale & 0.027*wicked + 0.026*spoon + 0.015*buffet + 0.014*dishes + 0.012*cosmopolitan + 0.011*stone + 0.011*mac + 0.011*marrow + 0.011*bone + 0.010*gelato \\ \hline

1 & Steak \& Eggs & 0.049*steak + 0.014*french + 0.012*eggs + 0.011*breakfast + 0.010*soup + 0.009*fries + 0.006*toast + 0.006*onion + 0.006*seated + 0.006*bread \\  \hline

2 & Tacos & 0.043*tacos + 0.037*taco + 0.026*asada + 0.024*carne + 0.024*mexican + 0.019*burrito + 0.010*salsa + 0.009*fries + 0.009*beans + 0.008*roberto's \\ \hline

3 & Mexican & 0.051*tacos + 0.025*pork + 0.021*corn + 0.014*grill + 0.013*bobby + 0.012*taco + 0.012*mesa + 0.009*tamale + 0.008*pastor + 0.007*chile \\ \hline

4 & Pho & 0.078*pho + 0.018*vietnamese + 0.014*rolls + 0.014*broth + 0.011*pork + 0.010*spring + 0.010*soup + 0.009*bowl + 0.008*rice + 0.008*noodles \\ \hline

5 & Seafood/Buffet & 0.057*buffet + 0.016*crab + 0.013*legs + 0.013*buffets + 0.012*line + 0.012*selection + 0.011*dessert + 0.010*desserts + 0.010*wynn + 0.008*bellagio \\ \hline

6 & Sports Bar & 0.031*beer + 0.010*beers + 0.009*wings + 0.007*bartender + 0.007*waitress + 0.007*server + 0.007*pub + 0.007*chips + 0.006*selection + 0.005*game \\ \hline

7 & Nightlife & 0.013*music + 0.012*fun + 0.009*club + 0.007*cool + 0.006*party + 0.006*lounge + 0.005*group + 0.005*floor + 0.004*dance + 0.004*girls \\ \hline 

8 & Brunch & 0.044*breakfast + 0.019*eggs + 0.017*pancakes + 0.010*toast + 0.010*coffee + 0.008*egg + 0.008*omelet + 0.007*bacon + 0.007*potatoes + 0.007*hash \\ \hline 

9 & Ramen & 0.066*ramen + 0.021*broth + 0.017*pork + 0.017*noodles + 0.016*bowl + 0.014*japanese + 0.011*miso + 0.010*rice + 0.007*egg + 0.007*soup \\ \hline 
 
10 & Chinese & 0.048*rice + 0.017*chinese + 0.011*egg + 0.011*filipino + 0.011*delivery + 0.010*soup + 0.009*shrimp + 0.008*teriyaki + 0.008*panda + 0.007*adobo \\ \hline 

11 & Seafood & 0.043*crab + 0.033*lobster + 0.029*shrimp + 0.025*seafood + 0.015*crawfish + 0.012*juicy + 0.009*cajun + 0.009*oysters + 0.008*n + 0.008*fries \\ \hline 

12 & Dim Sum & 0.042*dim + 0.042*sum + 0.015*pork + 0.014*pot + 0.011*chinese + 0.009*dumplings + 0.009*dishes + 0.009*filipino + 0.009*bao + 0.007*soup \\ \hline 

13 & Vegan/Healthy & 0.030*vegan + 0.018*healthy + 0.009*vegetarian + 0.009*wrap + 0.009*options + 0.007*store + 0.007*smoothie + 0.006*veggie + 0.005*juice + 0.005*raw \\ \hline 

14 & Tapas & 0.041*tapas + 0.020*sangria + 0.013*dates + 0.012*firefly + 0.011*stuffed + 0.011*dishes + 0.011*paella + 0.009*dish + 0.009*wrapped + 0.008*bacon \\ \hline 

15 & Upscale & 0.014*view + 0.010*wine + 0.010*bellagio + 0.008*bread + 0.008*patio + 0.007*dish + 0.006*waiter + 0.006*dessert + 0.006*fountains + 0.005*french \\ \hline 

16 & Buffet & 0.075*buffet + 0.018*buffets + 0.011*line + 0.011*selection + 0.009*breakfast + 0.009*station + 0.007*dessert + 0.007*seafood + 0.006*variety + 0.006*crab \\ \hline 

17 & Indian & 0.048*indian + 0.022*buffet + 0.020*naan + 0.013*lamb + 0.012*dishes + 0.011*rice + 0.010*masala + 0.010*curry + 0.010*garlic + 0.009*dish \\ \hline 

18 & Burgers & 0.091*burger + 0.047*fries + 0.034*burgers + 0.007*bacon + 0.007*shake + 0.006*bun + 0.006*onion + 0.005*patty + 0.005*in-n-out + 0.005*onions \\ \hline 

19 & Greek & 0.032*greek + 0.026*gyro + 0.025*pita + 0.024*hummus + 0.014*mediterranean + 0.013*lamb + 0.013*rice + 0.013*fries + 0.012*kabob + 0.011*bread \\ \hline 

20 & Fish \& Chips & 0.034*fish + 0.012*shrimp + 0.007*grilled + 0.007*dish + 0.006*server + 0.006*chips + 0.006*appetizer + 0.005*dessert + 0.005*coconut + 0.005*mahi \\ \hline 

21 & Street Vendors & 0.034*fries + 0.024*dog + 0.013*dogs + 0.010*truck + 0.009*chili + 0.006*chicago + 0.006*strips + 0.005*drive + 0.005*fingers + 0.005*wings \\ \hline 

22 & Korean BBQ & 0.038*korean + 0.018*hawaiian + 0.017*rice + 0.015*bbq + 0.011*pork + 0.009*bulgogi + 0.009*kimchi + 0.009*soup + 0.009*poke + 0.009*dishes \\ \hline 

23 & Mexican/Bar & 0.028*mexican + 0.024*salsa + 0.024*chips + 0.014*tacos + 0.010*margaritas + 0.010*margarita + 0.010*guacamole + 0.009*beans + 0.007*rice + 0.007*taco \\ \hline 

24 & Italian & 0.019*pasta + 0.018*italian + 0.012*bread + 0.007*server + 0.007*dish + 0.006*waiter + 0.006*wine + 0.006*garlic + 0.005*meatballs + 0.005*spaghetti \\ \hline 

25 & Pizza & 0.121*pizza + 0.016*crust + 0.010*slice + 0.009*pizzas + 0.008*pepperoni + 0.007*garlic + 0.007*toppings + 0.006*slices + 0.006*pie + 0.005*sausage \\ \hline 

26 & BBQ & 0.043*bbq + 0.026*ribs + 0.019*mac + 0.018*pork + 0.013*brisket + 0.011*sides + 0.010*pulled + 0.010*beans + 0.007*potato + 0.007*tender \\ \hline 

27 & Burgers & 0.063*burger + 0.028*fries + 0.018*burgers + 0.009*frozen + 0.007*potato + 0.006*onion + 0.006*truffle + 0.005*shake + 0.005*kobe + 0.005*bacon \\ \hline 

28 & Thai & 0.074*thai + 0.022*pad + 0.017*curry + 0.014*rice + 0.011*dish + 0.010*dishes + 0.009*soup + 0.008*noodles + 0.007*tom + 0.005*duck \\ \hline 

29 & Waffles/Brunch & 0.025*waffles + 0.017*portions + 0.017*hash + 0.015*breakfast + 0.014*bacon + 0.011*eggs + 0.011*waffle + 0.009*potatoes + 0.008*sage + 0.007*benedict \\ \hline 

30 & Bad Service & 0.008*manager + 0.008*cafe + 0.007*worst + 0.006*slow + 0.006*horrible + 0.005*server + 0.005*airport + 0.005*waitress + 0.005*rude + 0.005*sandwich \\ \hline 

31 & Luxe & 0.015*cheesecake + 0.012*factory + 0.012*grand + 0.009*sliders + 0.008*venetian + 0.008*cafe + 0.008*breakfast + 0.008*sandwich + 0.007*palazzo + 0.007*lux \\ \hline 
 
32 & Dessert & 0.054*chocolate + 0.021*cake + 0.017*bouchon + 0.015*cream + 0.012*crepe + 0.011*french + 0.011*pastries + 0.010*bakery + 0.009*dessert + 0.008*coffee \\ \hline 

33 & Sushi & 0.033*sushi + 0.014*roll + 0.011*sashimi + 0.011*japanese + 0.011*tuna + 0.009*fish + 0.008*sake + 0.008*rolls + 0.007*dishes + 0.007*miso \\ \hline 

34 & Deli & 0.069*sandwich + 0.031*sandwiches + 0.019*bread + 0.010*turkey + 0.008*deli + 0.006*pastrami + 0.006*soup + 0.006*line + 0.005*earl + 0.004*sub \\ \hline 

35 & Asian/Authentic & 0.048*curry + 0.018*rice + 0.015*soup + 0.012*owner + 0.010*pork + 0.008*dish + 0.008*katsu + 0.006*authentic + 0.006*spice + 0.006*dishes \\ \hline 

36 & Burritos & 0.032*burrito + 0.018*rice + 0.018*chipotle + 0.017*tacos + 0.013*bowl + 0.013*salsa + 0.010*fusion + 0.009*beans + 0.009*steak + 0.008*chips \\ \hline 

37 & Steakhouse & 0.028*steak + 0.009*filet + 0.007*lobster + 0.007*steaks + 0.006*sides + 0.006*waiter + 0.006*wine + 0.006*bread + 0.006*dessert + 0.006*steakhouse \\ \hline 

38 & Exotic American & 0.013*pork + 0.010*burger + 0.009*belly + 0.008*fries + 0.008*egg + 0.008*truffle + 0.007*oxtail + 0.007*bachi + 0.007*flavors + 0.006*dish \\ \hline 

39 & Wine/Upscale & 0.021*wine + 0.008*bread + 0.007*steak + 0.006*waiter + 0.005*server + 0.005*dessert + 0.005*chef + 0.004*pasta + 0.004*dish + 0.004*wonderful \\ \hline 

40 & Cafe  & 0.050*coffee + 0.039*tea + 0.021*ice + 0.012*boba + 0.010*milk + 0.008*hookah + 0.008*cup + 0.007*iced + 0.007*shop + 0.007*latte \\ \hline 

41 & Upscale & 0.010*wine + 0.009*dish + 0.009*dessert + 0.008*foie + 0.008*gras + 0.007*lobster + 0.007*tasting + 0.006*bread + 0.005*chocolate + 0.005*dishes \\ \hline 

42 & Sushi & 0.079*sushi + 0.027*rolls + 0.025*roll + 0.018*ayce + 0.014*fish + 0.009*rice + 0.009*tuna + 0.008*nigiri + 0.007*salmon + 0.006*cream \\ \hline 

43 & Soup & 0.160*soup + 0.032*noodle + 0.023*donuts + 0.018*island + 0.017*treasure + 0.017*ti + 0.015*noodles + 0.012*excalibur + 0.011*donut + 0.011*soups \\ \hline 

44 & Oysters & 0.040*oysters + 0.035*roast + 0.029*pan + 0.026*oyster + 0.015*chowder + 0.014*gumbo + 0.014*clam + 0.013*seafood + 0.012*line + 0.011*palace \\ \hline 

45 & Casino/Hotel & 0.027*hotel + 0.022*casino + 0.020*room + 0.010*stay + 0.009*downtown + 0.006*staying + 0.006*pool + 0.005*street + 0.005*stayed + 0.005*fremont \\ \hline 

46 & Noodles & 0.029*noodles + 0.027*chinese + 0.024*noodle + 0.019*soup + 0.019*rice + 0.013*asian + 0.012*duck + 0.011*dish + 0.010*dishes + 0.009*pork \\ \hline 

47 & Chinese & 0.030*chinese + 0.017*shrimp + 0.013*rice + 0.010*pork + 0.010*dishes + 0.010*dish + 0.008*soup + 0.007*fish + 0.006*pepper + 0.006*crispy \\ \hline 

48 & Buffet & 0.022*line + 0.019*rio + 0.018*buffet + 0.017*bacchanal + 0.015*tots + 0.015*seafood + 0.011*station + 0.011*caesar's + 0.010*caesars + 0.010*oysters \\ \hline 

49 & Prime Rib & 0.164*rib + 0.142*prime + 0.029*cut + 0.026*potatoes + 0.022*mashed + 0.014*potato + 0.013*baked + 0.011*medium + 0.011*lawry's + 0.010*rare \\ \hline 
    \end{longtable}
\end{center}

\begin{figure}
    \caption{Labeling Performed on Other Cities}
    \centering
    \subfloat[Phoenix]{{\includegraphics[width=0.4\textwidth]{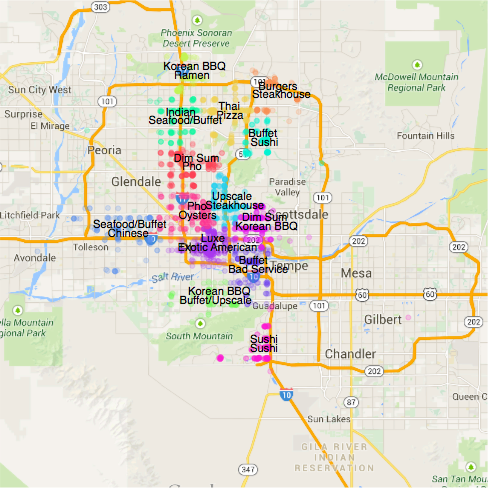} }}
    \qquad
    \subfloat[Edinburgh]{{\includegraphics[width=0.4\textwidth]{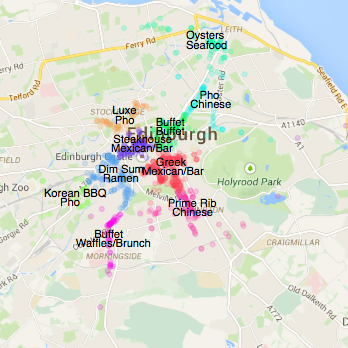} }}
    \label{fig:example}
\end{figure}

\newpage

\begin{center}
More Advanced Calculation of S
\end{center}

\begin{gather}
Dist(CR,C) < Dist(CR, I)\nonumber\\
\sqrt{(0.75)^2 + S*(0.0)^2} < \sqrt{(0.25)^2 + 2*(S*0.5)^2 + 10*(S*0.1)^2}\nonumber\\
0.75 < \sqrt{(0.25)^2 + 2*(0.5*S)^2 + 10*S*(0.1*S)^2}\nonumber\\
0.5625 < 0.0625 + 0.6S^2\nonumber\\
\sqrt{\frac{5}{6}} < S\nonumber\\
S > 0.913
\end{gather}

\end{document}